\documentclass[journal]{IEEEtran}
\usepackage{amsmath,amsfonts}
\usepackage{algorithm}
\usepackage[noend]{algpseudocode}
\usepackage{array}
\usepackage[caption=false,font=normalsize,labelfont=sf,textfont=sf]{subfig}
\usepackage{caption}
\usepackage{textcomp}
\usepackage{stfloats}
\usepackage{url}
\usepackage{verbatim}
\usepackage{graphicx}
\usepackage{cite}
\usepackage{amssymb}
\usepackage{diagbox}
\usepackage{subfig}
\usepackage{colortbl}
\usepackage[table]{xcolor}
\usepackage{tabularx}
\usepackage{multirow}
\usepackage{makecell}
\usepackage{booktabs}

\begin{document}

\title{Leveraging Generalizability of Image-to-Image Translation for Enhanced Adversarial Defense}

\author{Haibo Zhang,~\IEEEmembership{Member,~IEEE},
    Zhihua Yao,
    Kouichi Sakurai,~\IEEEmembership{Member,~IEEE},
    and Takeshi Saitoh,~\IEEEmembership{Member,~IEEE} 
\thanks{Haibo Zhang is with the Department of Artificial Intelligence, Faculty of Computer Science and Systems Engineering, Kyushu Institute of Technology, Japan (e-mail: haiboz0105@gmail.com).}
\thanks{Zhihua Yao is with the Faculty of Economics and Business Administration, The University of Kitakyushu, Japan (e-mail: zhihuayao@alumni.usc.edu).}
\thanks{Kouichi Sakurai is with the Department of Informatics, Faculty of Information Science and Electrical Engineering, Kyushu University, Japan (e-mail: sakurai@inf.kyushu-u.ac.jp).}
\thanks{Takeshi Saitoh is with the Department of Artificial Intelligence, Faculty of Computer Science and Systems Engineering, Kyushu Institute of Technology, Japan (e-mail: saitoh@ai.kyutech.ac.jp).}}



\maketitle

\begin{abstract}
In the rapidly evolving field of artificial intelligence, machine learning emerges as a key technology characterized by its vast potential and inherent risks. The stability and reliability of these models are important, as they are frequent targets of security threats. Adversarial attacks, first rigorously defined by Ian Goodfellow et al. in 2013, highlight a critical vulnerability: they can trick machine learning models into making incorrect predictions by applying nearly invisible perturbations to images. Although many studies have focused on constructing sophisticated defensive mechanisms to mitigate such attacks, they often overlook the substantial time and computational costs of training and maintaining these models. Ideally, a defense method should be able to generalize across various, even unseen, adversarial attacks with minimal overhead. Building on our previous work on image-to-image translation-based defenses, this study introduces an improved model that incorporates residual blocks to enhance generalizability. The proposed method requires training only a single model, effectively defends against diverse attack types, and is well-transferable between different target models. Experiments show that our model can restore the classification accuracy from near zero to an average of 72\% while maintaining competitive performance compared to state-of-the-art methods. Significantly, our model operates more efficiently, reducing the time needed to process individual images and speeding up the training process to achieve faster convergence.  Robustness tests further confirm stable performance under varying attack strengths, demonstrating the model’s practical value in real-world adversarial settings.
\end{abstract}

\section{Introduction and Background}

Deep Neural Networks (DNNs) represent a foundation in the landscape of deep learning models, with broad applicability in diverse image recognition tasks, including object detection, facial recognition, and autonomous driving. Despite the success, extensive research indicates that these models are highly susceptible to adversarial attacks, as substantiated by seminal studies such as those conducted by Goodfellow et al. \cite{goodfellow2014explaining}. These adversarial attacks involve subtle modifications to the images that are meticulously crafted and are sufficient to mislead the model into making classification errors.

\begin{figure}
\centering
\includegraphics[width=0.48\textwidth]{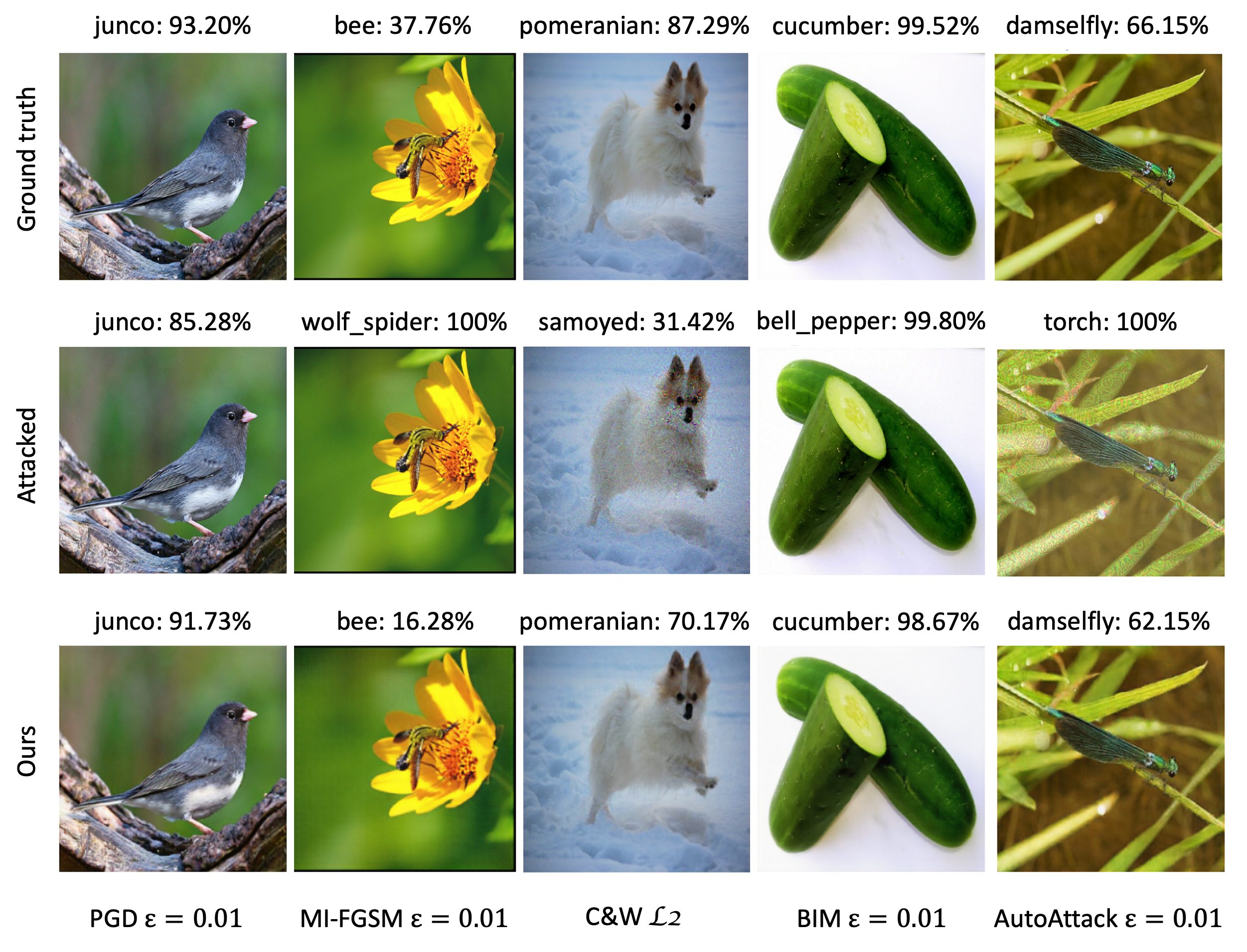}
\caption{Adversarial examples of FGSM attack, PGD attack, C\&W attack and AutoAttack.} 
\end{figure}

The vulnerability of machine learning models to such attacks has significant implications, particularly in critical applications such as facial recognition and autonomous driving. Therefore, it is important to develop and deploy strong defenses against adversarial attacks to maintain the security of machine learning systems, particularly in safety-critical settings. The importance of this need is highlighted by studies from researchers such as Xu et al. \cite{xu2020adversarial}, who stress that reinforcing these models against threats is crucial for their secure and extensive implementation.

Defending against adversarial attacks is comparable to the way antivirus software operates. Due to the continued advancement and variety of adversarial techniques, having the ability to generalize across a wide range of attacks is crucial to evaluating defensive strategies. Resource allocation is also essential, covering both training expenses and the time required for computation. Training models customized for specific attacks and updating the defense database is resource-intensive and costly. In an ideal scenario, a defense mechanism should utilize a flexible model that demonstrates significant generalizability and robustness when faced with known and unexpected attacks \cite{zhang2024meta,lau2023interpolated,yin2023generalizable}.

Our study investigates a defense method that employs image-to-image translation techniques within this context. To evaluate the generalizability and effectiveness of this approach, we employed a broad range of experimental methods, which included tests across different attack types, evaluations with various models, trials with different datasets, and thorough verifications of robustness. To assess the defense strategy's ability to resist evolving threats, robustness testing played a crucial role by simulating situations where attackers persistently adapted their methods. This method notably strengthened the model's defensive stance by elevating its capability to predict and neutralize existing and developing attack strategies. Our results, as shown in Fig. 1, demonstrate that the image-to-image translation defense method maintains its effectiveness in various attack situations and outperforms conventional defense mechanisms for adaptability and durability.

Consequently, this research expands the potential for adversarial defense strategies and presents methods for developing secure future machine learning systems. By utilizing the adaptability of image-to-image translation technology, we showcase a flexible and efficient defense approach that adjusts to different adversarial contexts, all while preserving strong protective abilities, thus playing a crucial role in the progress of the machine learning discipline.

\subsection{Research Motivation}
Though significant research has somewhat reduced the effects of adversarial attacks on machine learning models, the effectiveness of these defenses against new or unforeseen attacks is still questionable. In practical situations, adversaries can use various strategies and methods, which require a defense model capable of adjusting to and resisting numerous attacks. Therefore, exploring the generalizability of existing defense mechanisms across different attack modalities is important to improve the security and robustness of models.

In earlier studies \cite{zhang2021conditional,zhang2023eliminating}, we illustrated the strong efficacy of the Generative Adversarial Network \cite{goodfellow2020generative}, particularly concerning image-to-image translation methods in counteracting a particular class of adversarial attacks. This paper seeks to expand on this work by investigating the generalizability of these defense strategies. We pose several research questions: 

\begin{enumerate}
    \item \textbf{Q1: Is it possible for a model, trained on a particular set of adversarial attack data, to successfully defend against other types of adversarial threats?}
    
    To outline the problem, consider a model \( M \) that has been trained on a dataset \( D_{train} \) containing samples \( (x, y) \) altered by a particular adversarial attack \( \mathcal{A} \). The objective is to assess if this model retains its accuracy when tested with a dataset \( D_{test} \) affected by distinct adversarial attacks \( \mathcal{B} \).

    The general form of a model's objective in adversarial training can be modeled as:
    \begin{equation}
        \min_{\theta} \mathbb{E}_{(x, y) \sim D_{train}} \left[ \max_{\delta \in \mathcal{A}(x)} \mathcal{L}(f(x + \delta; \theta), y) \right] , 
    \end{equation}
    
    where:
    \begin{itemize}
      \item \( \theta \) represents the model parameters.
      \item \( \delta \) represents the perturbation added to the input \( x \) under attack type \( \mathcal{A} \).
      \item \( \mathcal{L} \) is the loss function measuring the discrepancy between the model's prediction on the adversarial example and the true label \( y \).
    \end{itemize}
    
    To test the generalizability of the model to different attacks, we can define a generalizability metric \( G \) as:
    \begin{equation}
    G = 1 - \frac{\text{Accuracy on } D_{test} \text{ with attack } \mathcal{B}}{\text{Accuracy on } D_{test} \text{ with attack } \mathcal{A}}
    \end{equation}
    
    A low value of \( G \) indicates better generalizability, suggesting that the model's defense capabilities generalize well across different types of attacks. 

    \item \textbf{Q2: Is it possible for a model trained on a diverse set of attack datasets to exhibit robust defense against novel or previously unencountered attacks?} 

    We need a unified model \( M' \) that can learn to defend against a diverse range of adversarial attacks, ideally performing as well as or better than multiple specialized models. 

    Let \( \mathcal{A}_i \) represent the adversarial attack type \( i \), where \( i \in \{1, 2, \dots, n\} \). Assume \( D_i \) is the dataset containing examples that have been perturbed by attack \( \mathcal{A}_i \).
    
    Traditionally, a model \( M_i \) is trained specifically for each attack type \( \mathcal{A}_i \):
    \begin{equation}
         M_i = \text{train}(D_i) 
    \end{equation}
    where "train" denotes the training process tailored to optimize performance against \( \mathcal{A}_i \).
    
    We aim to train a single model \( M' \) using a combined dataset from all attack types. Let \( D \) be the aggregated dataset: $ D = \bigcup_{i=1}^n D_i $.
    The training objective for \( M' \) is to minimize the expected loss over \( D \), considering all attack types:
    \begin{equation}
        M' = \text{train}(D)
    \end{equation}
    
    The loss function \( \mathcal{L} \) for \( M' \) needs to capture the performance across all types of attacks effectively. The model \( M' \) is optimized by:
    \begin{equation}
        \min_{\theta} \sum_{i=1}^n \mathbb{E}_{(x, y) \in D_i} \left[ \mathcal{L}(f(x; \theta), y) \right] , 
    \end{equation}
    where \( \theta \) represents the parameters of \( M' \), and \( f(x; \theta) \) is the prediction function of \( M' \).
        
    \item \textbf{Q3: Are defense strategies designed for a specific model equally effective against attacks aimed at other models?}

    Let \( M_a \) and \( M_b \) be two different models subjected to the same type of adversarial attack \( \mathcal{A} \). A defense model \( M_D \) is trained using data \( D_{M_a} \) generated by attacking \( M_a \) with \( \mathcal{A} \). The objective is to mitigate the effects of \( \mathcal{A} \), aiming to restore inputs or predict correct outputs despite adversarial modifications.
    
    The efficacy of \( M_D \), originally trained on \( D_{M_a} \), is tested on a new dataset \( D_{M_b} \) generated by the same attack \( \mathcal{A} \) on \( M_b \).

\end{enumerate}

We aim to systematically validate experiments to investigate and establish more broadly applicable defense strategies. This effort will further the progress of adversarial attack defense technologies, providing strong support for research and applications within related domains.

\subsection{Contributions}

Previous investigations by the authors have shown that image-to-image translation technology is a viable means of defending against adversarial attacks. Expanding on this foundation, the current study delves deeper into and confirms the generalizability of adversarial defense strategies that utilize image-to-image translation techniques. The primary contributions of this study include:

\begin{itemize} 
\item This study developed an improved model that combines image-to-image translation techniques with residual blocks, utilizing a composite dataset that includes a variety of adversarial attacks. 
\item This study validated our proposed model on four different levels of datasets (MNIST, F-MNIST, CIFAR-10, and ImageNet). The model demonstrated comparability with leading methods across key performance evaluation metrics, including image recognition accuracy, image quality (measured by the Peak Signal-to-Noise Ratio, PSNR), training epochs, and processing speed. 
\item Furthermore, this study trained models specifically against single-type attack datasets and successfully applied these models to defend against other types of adversarial attacks, further proving the high generalizability of our approach. 
\item This study further investigated how well this defense strategy can be applied to various target models.
\item Subsequently, this study conducted extensive robustness analyzes on the proposed model, confirming its efficacy in practical applications.
\end{itemize}

\section{Related Works}
This section will comprehensively discuss related work on adversarial attacks, some primary defense methods, and image-to-image translation technology.

\subsection{Adversarial Attacks}
Adversarial attacks involve techniques that apply minor adjustments to the input data provided to a model. These changes are often barely noticeable to human eyes, yet they can potentially cause the model to incorrectly classify the input \cite{huang2022masked,zhao2022clpa}. These attacks are mathematically modeled using an original image \(x\), correctly classified in the category \(y\). The objective involves a classifier \(f\) (such as a deep neural network), which maps an input image \(x\) to a predicted category \( \hat{y} \). The objective of the attacker is to find a minimal perturbation \( \delta \) such that, when incorporated into \( x \), it causes the classifier to output an incorrect prediction \( \hat{y}' \), with \( \hat{y}' \neq y \), as the Fig. 2 shown. The perturbation \( \delta \) is restricted to being invisible to human observers, often regulated by norms like \( L_0 \), \( L_2 \), and \( L_{\infty} \). The creation of these adversarial examples can be framed as an optimization problem: \(\min_{\delta} \| \delta \| \) subject to \( f(x + \delta) = \hat{y}' \), or \(\min_{\delta} L(f(x + \delta), \hat{y}') + \lambda \| \delta \| \), where \( L \) is a loss function measuring the deviation between \( f(x + \delta) \) and the target category \( \hat{y}' \), and \( \lambda \) serves as a regularization parameter to balance the magnitude of perturbation against the need for misclassification.

\begin{figure}[!ht]
    \centering
    \includegraphics[width=0.9\linewidth]{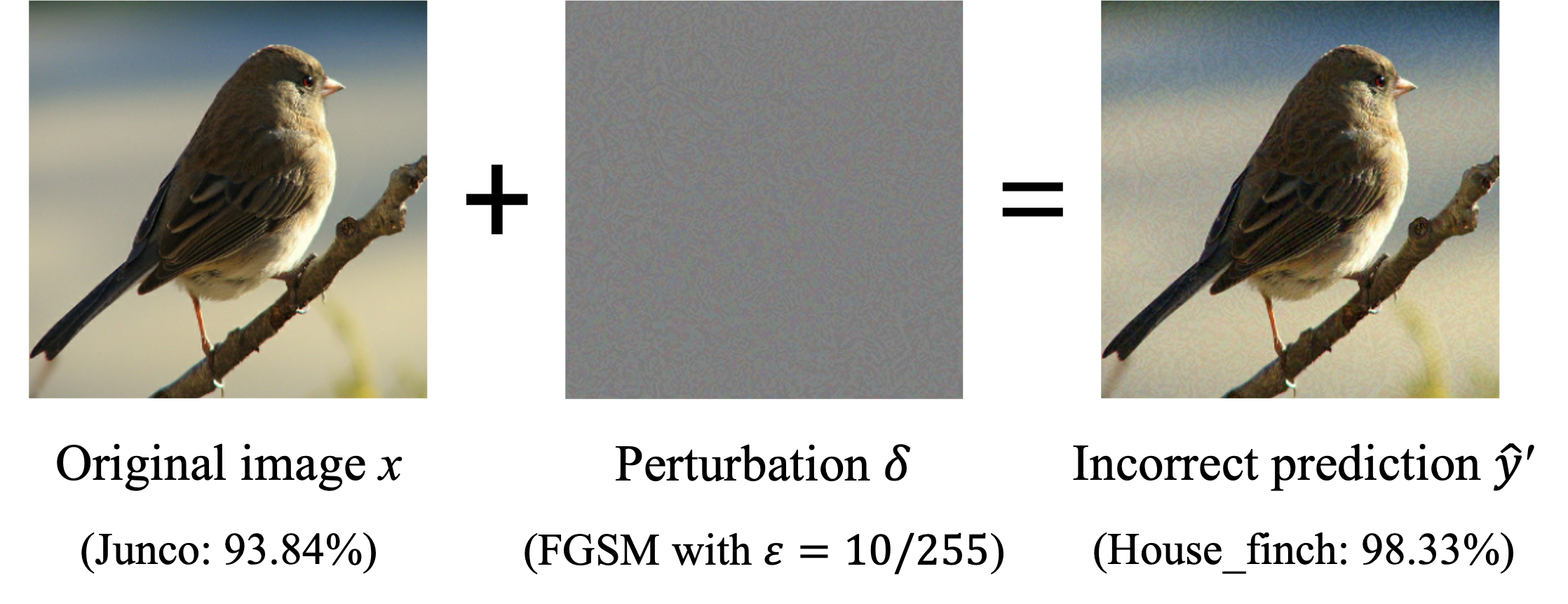}
    \caption{An adversarial example under the FGSM attack with $\epsilon = 10/255$.}
\end{figure}

In this research, our primary objective is to delve into the vulnerabilities of the target model and improve its robustness, concentrating solely on white-box attacks to examine defense strategies. Some representative attacks dealt with in this study including Fast Gradient Sign Method (FGSM)\cite{goodfellow2014explaining}, Basic Iterative Method (BIM)\cite{kurakin2018adversarial}, Projected Gradient Descent (PGD)\cite{madry2017towards}, Carlini \& Wagner (C\&W)\cite{carlini2017towards}, Momentum Iterative Fast Gradient Sign Method (MI-FGSM)\cite{dong2018boosting}, DeepFool\cite{moosavi2016deepfool}, and AutoAttack\cite{croce2020reliable}. 

\noindent
\textbf{Fast Gradient Sign Method (FGSM)}
    FGSM exploits the gradients of the loss for the input data to create adversarial examples by adding a perturbation determined by the sign of the gradient of the data. The adversarial example $x'$ is constructed as follows:
    \begin{equation}
         x' = x + \epsilon \cdot sign(\nabla_x J(\theta, x, y)) 
    \end{equation}   
    where $J(\theta, x, y)$ is the loss function used by the network, $\nabla_x$ denotes the gradient with respect to input $x$, $y$ is the correct label for $x$, $\theta$ represents the model parameters, and $\epsilon$ is a small constant.
    
\noindent
\textbf{Basic Iterative Method (BIM) \& Projected Gradient Descent (PGD)}
    Evolving from FGSM, BIM and PGD implement iterative assaults. Both methods execute iterative gradient steps using a diminutive step size, accompanied by clipping, to guarantee the adversarial examples remain within a stipulated vicinity.
    
    \begin{equation}
    x^{(N+1)} = {Clip}_{x,\epsilon}(x^{(N)} + \alpha \cdot sign(\nabla_x J(\theta, x^{(N)}, y)))
    \end{equation}
\noindent
\textbf{Momentum Iterative Fast Gradient Sign Method (MI-FGSM)}
    Building upon FGSM, MI-FGSM introduces a momentum component to navigate better and escape unfavorable local maxima.
    
    \begin{equation}
    g^{(N+1)} = \mu \cdot g^{(N)} + \frac{\nabla_x J(\theta, x^{(N)}, y)}{\|\nabla_x J(\theta, x^{(N)}, y)\|_1},
    \end{equation}
    
    \begin{equation}
    x^{(N+1)} = {Clip}_{x,\epsilon}(x^{(N)} + \alpha \cdot sign(g^{(N+1)}))
    \end{equation}
    
    where $N$ is the iteration number, $\mu$ is the decay factor of the momentum.
    
\noindent
\textbf{Carlini \& Wagner (C\&W)}
    The C\&W attack formulates the problem as an optimization problem. It aims to find the smallest perturbation that results in misclassification. The attack optimizes both the perturbation and the confidence of the misclassification. The optimization objective is:
    
    \begin{equation}
    \min_{\delta} \| \delta \|_2^2 + c \cdot f(x + \delta)
    \end{equation}
    
    where $f(x + \delta)$ is the classification loss, and $c$ is a constant to control the trade-off between the perturbation size and the misclassification rate.

\noindent
\textbf{DeepFool}
    DeepFool operates by iteratively linearizing the classifier at the current point and moving toward the resulting decision boundary. This results in potentially smaller perturbations than other methods. The perturbation at each step can be computed as follows:
    
    \begin{equation}
    r^{(N)} = -\frac{f_k(x^{(N)})}{\|\nabla f_k(x^{(N)})\|^2}\nabla f_k(x^{(N)})
    \end{equation}
    
    where $f_k$ is the k-th class score. The updated point is $x^{(N+1)} = x^{(N)} + r^{(N)}$.

\noindent
\textbf{AutoAttack}
    AutoAttack is a comprehensive automated toolkit combining several white-box and black-box adversarial attack methods for machine learning models. Its primary goal is to thoroughly and reliably assess the resilience of machine learning models against adversarial threats. AutoAttack encompasses four distinct attacks: 1) Auto-PGD: an advanced adaptation of the PGD attack. 2) APGD-DLR: a variant of Auto-PGD utilizing a decision-based loss function with randomized smoothing techniques. 3) FAB (Fast Adaptive Boundary): an optimized attack method for crafting adversarial perturbations utilizing L1 and L2 norms. 4) Square Attack: a black-box score-based technique that alters an image's local section of contiguous pixels.

\subsection{Defense on Adversarial Attacks}
Researchers are diligently working on various defense strategies to mitigate the detrimental effects of adversarial attacks. These methods are generally classified into two types: reactive defenses, which focus on identifying and counteracting adversarial examples, and proactive defenses, designed to prepare models to deal with such perturbations \cite{bountakas2023defense}. Although many of these defense systems appear initially effective, they often fall short when faced with more advanced and adaptable adversarial tactics \cite{ho2022disco, mo2022adversarial}. This highlights the need for a thorough evaluation and stringent benchmarking of defense strategies.

\subsubsection{\textbf{On Adversarial Training}}
Adversarial training is recognized as a proactive strategy for defense \cite{li2022new}. The groundbreaking research of Goodfellow et al. \cite{goodfellow2014explaining} and subsequent contributions by Madry et al. \cite{madry2017towards} emphasized the importance of integrating adversarial instances into training sets to strengthen the robustness of the model. Based on these principles, Kurakin et al. \cite{kurakin2016adversarial} suggested using clean and adversarial images within training batches to fortify network resilience further. Tramèr et al. \cite{tramer2017ensemble} advanced this idea by incorporating perturbations from multiple models, thus enhancing the overall resistance of the training process.

The advent of Adversarial Logit Pairing (ALP) by Kannan et al. \cite{kannan2018adversarial} marked a significant innovation by aligning the logit outputs of both clean and adversarial images, thereby promoting the assimilation of features from clean images. This proved particularly effective on intricate datasets like ImageNet. Building on ALP, Xie et al. \cite{xie2019feature} developed a method that incorporates a denoising framework into the high-level feature maps, enhancing the distinction of critical image features. This approach has shown increased robustness, especially against PGD attacks.

\subsubsection{\textbf{On Pre-processing Images}}
Pre-processing defenses concentrate on modifying or cleansing input images prior to model processing. Liao et al. \cite{liao2018defense} introduced the High-Level Representation-Guided Denoiser (HGD), designed to correct distorted inputs by ensuring that their high-level features resemble those of authentic images. In a parallel effort, Song et al. \cite{song2017pixeldefend} formulated PixelDefend, which purifies images by altering their statistical properties to align with clean training data, thereby reducing the effects of adversarial noise.

Meng et al. \cite{meng2017magnet} proposed MagNet, which successfully integrates detection and restoration networks to combat black-box attack scenarios. In an innovative application of generative models, Samangouei et al. \cite{samangouei2018defense} unveiled Defense-GAN, which employs generative adversarial networks to rebuild inputs and eliminate adversarial alterations before they impact the target model.

Expanding the defensive toolkit, Zhang et al. \cite{zhang2021defense} devised a reconstruction method that improves defenses by reshaping adversarial images to more closely emulate the distribution of clean data, with promising results on high-resolution datasets such as ImageNet ILSVRC2012. This approach is complemented by the randomization method from Xie et al. \cite{xie2017mitigating}, which disrupts adversarial perturbations by randomizing neural network layers, and the Pixel Deflection method by Prakash et al. \cite{prakash2018deflecting}, which confuses adversarial mechanisms by deflecting pixels within images.

Furthermore, Mustafa et al. \cite{mustafa2019image} presented the super-resolution approach, which utilizes enhancement methods to improve image resilience to attacks, acting as a strong pre-processing defense. Similarly, Donoho's well-known wavelet denoising method \cite{donoho1995noising} uses wavelet transforms to eliminate noise, effectively reducing the impact of adversarial disturbances.

\subsubsection{\textbf{Alternative Defense Mechanisms}}
Papernot et al. \cite{papernot2016distillation} investigated defensive distillation, which trains the target model using soft labels from a surrogate model. This approach significantly lowers the success rates of attacks; however, Carlini \& Wagner \cite{carlini2017adversarial} have identified certain weaknesses.

The concept of feature squeezing introduced by Xu et al. \cite{xu2017feature} focuses on diminishing the adversarial search space by constraining the granularity of input features, thereby diminishing the impact of small perturbations. Meanwhile, Cohen et al. \cite{cohen2019certified} investigated randomized smoothing, which involves adding noise to input samples as a method to counteract adversarial modifications, offering a certifiable safeguard in certain contexts. Moreover, adaptive defensive dropout techniques adjust the dropout rate in response to the anticipated probability of adversarial attacks, thus reinforcing model robustness as demonstrated in foundational research by Srivastava et al. \cite{srivastava2014dropout}.

These strategies underscore the intricate and ever-changing challenge of crafting effective defenses to counter adversarial attacks. They also highlight the necessity for ongoing innovation and thorough testing to stay abreast of the advancement of adversarial methods.

\subsection{Image-to-image Translation}
Image-to-image translation is a distinct area within the field of computer vision that aims to transform images from one domain into those of another. This conversion is mainly achieved through the use of Generative Adversarial Networks (GANs). Image-to-image translation represents a particular application of Conditional Generative Adversarial Networks (cGANs) \cite{mirza2014conditional}. Prominent approaches in this sphere include Pix2pix \cite{isola2017image} and StyleGAN \cite{karras2019style}, which act as key frameworks for carrying out image translations. These techniques have shown their adaptability and efficacy in various applications, such as converting satellite images into maps and altering daytime visuals to appear as nighttime scenes \cite{kurakin2018adversarial}.

\subsubsection{\textbf{Applicability on Image Reconstruction}}
The effectiveness of image-to-image translation methods in enhancing defenses against adversarial attacks is mainly due to their use of GANs. These networks are exceptional in producing high-quality images and provide an innovative approach to address adversarial perturbations. With GANs, models can be proficiently trained to identify and reconstruct clean images from those affected by adversarial noise. 

Furthermore, a significant advantage of GANs is their ability to handle complex mappings between distinct image domains expertly. This expertise allows for crafting customized defense strategies aimed at particular attack vectors. For example, specialized image translation models can be designed to identify and mitigate these disruptions based on specific perturbation patterns. This targeted strategy consequently helps shield downstream image classification or recognition models from adversarial influences, thus improving the overall resilience of the system.

\subsubsection{\textbf{Advantages}} 
Image-to-image translation techniques can provide notable benefits for analyzing adversarial attack defense methods, including increased model resilience, greater transparency in defense mechanisms, and enhanced interpretability. This method enables visualization of the alterations made to an image. 

Utilizing image-to-image translation methods during the pre-processing phase of the imaging pipeline greatly enhances the system's resilience. These methods improve security by preemptively eliminating possible adversarial disturbances. This strategy effectively "cleanses" images and greatly enhances the defense mechanism's interpretability, offering clearer insights into how adversarial perturbations are managed. Image-to-image translation methods are noted for their proficiency in learning and executing various image feature alterations. This enables them to effectively manage different categories of adversarial attacks and adapt to changing attack scenarios. As a result, these techniques show strong generalizability in varied and evolving adversarial contexts.

\section{Methods}
This section provides a detailed introduction to the image reconstruction defense method proposed in this study, which is based on image translation technology. 

\subsection{Model Design}

\begin{figure*}[ht]
\centering
\includegraphics[width=0.97\textwidth]{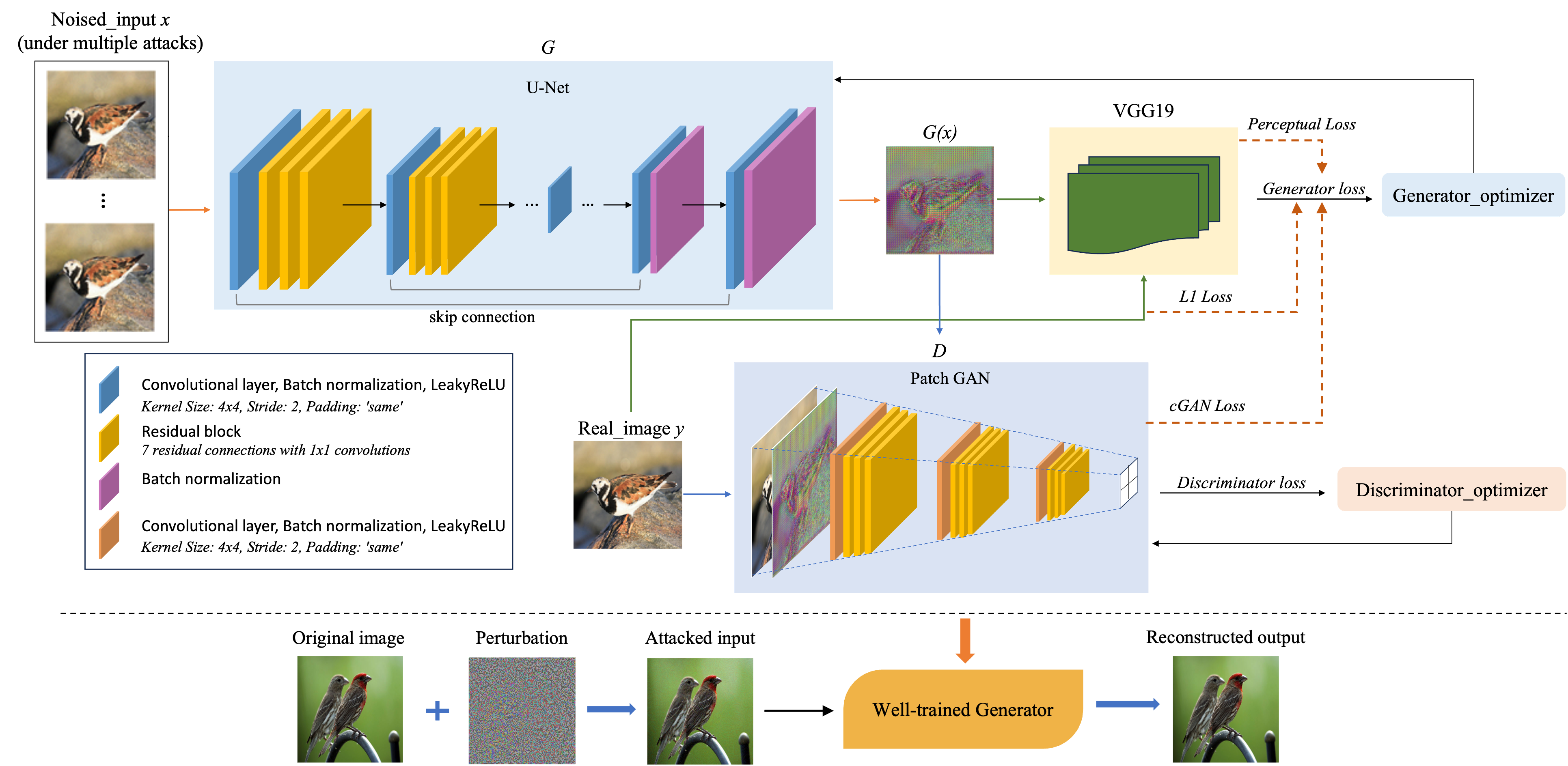}
\caption{Comprehensive architecture of the proposed image reconstruction method for defending against adversarial attacks.} 
\end{figure*}

The core structure of this approach is the conditional GAN framework, comprising a generator \(G\) and a discriminator \(D\), as illustrated in Fig. 3. Expanding on this basis, we utilize a generator that combines the U-Net architecture with Residual Blocks, a configuration influenced by the pix2pix model. This generator is specifically designed to handle the input noisy image \(x\). It refines and extracts features progressively through several residual blocks to generate an output \(G(x)\) that is clearer and bears a closer resemblance to the real image. In this framework, the generator's output is utilized not only for image generation but is also passed into a feature extractor that further examines the generated image’s features. These are then compared with the real image features \(y\) to provide feedback for training.

\begin{algorithm}
\caption{Model Description}
\begin{algorithmic}[1]
\State \textbf{Input:} Attacked image $I_{\text{att}}$
\State \textbf{Output:} Reconstructed image $I_{\text{rec}}$
\newline
\State Initialize U-Net, residual blocks $R\_b$, generator $G$, and discriminator $D$.
\State Load pre-trained VGG19 model for perceptual loss calculation.
\State $I_{\text{inter}} \gets$ Apply U-Net to $I_{\text{att}}$ for initial image enhancement.
\State $I_{\text{enhan}} \gets$ Apply $R\_b$ to $I_{\text{inter}}$ for further enhancement.
\State $I_{\text{fake}} \gets$ Generate a preliminary fake image using $G$ from $I_{\text{enhan}}$.

\Repeat
    \State \textbf{(a)} $L_{\text{content}} \gets$ Compute perceptual loss using VGG19 between $I_{\text{fake}}$ and $I_{\text{att}}$.
    \State \textbf{(b)} Optimize $G$ to minimize $L_{\text{content}}$.
    \State \textbf{(c)} $L_{\text{adv}} \gets$ Evaluate $D$ using $I_{\text{fake}}$ and $I_{\text{att}}$ to assess adversarial robustness.
    \State \textbf{(d)} Optimize $D$ to maximize $L_{\text{adv}}$.
    \State \textbf{(e)} Update $G$ based on $D$ feedback to enhance $I_{\text{fake}}$.
\Until{convergence}

\State $I_{\text{rec}} \gets I_{\text{fake}}$
\State \textbf{return} $I_{\text{rec}}$
\end{algorithmic}
\end{algorithm}

Algorithm 1 presents our proposed method for reconstructing images that have undergone adversarial attacks. The procedure starts with the input of an attacked image, $I_{\text{att}}$, and seeks to produce a reconstructed image, $I_{\text{rec}}$, as the output. Initially, key elements of the model are set up, including U-Net, residual blocks, a generator, and a discriminator, which are vital for improving and recovering the image. After this setup, the algorithm leverages a pre-trained VGG19 model to calculate perceptual loss, which is crucial for evaluating the similarity between the attacked and reconstructed images.

The image processing task begins by employing the U-Net on the attacked image to create an intermediate enhanced image $I_{\text{inter}}$. This intermediate image undergoes additional enhancement via residual blocks, producing $I_{\text{enhan}}$. The generator $G$ then utilizes $I_{\text{enhan}}$ to generate an initial fake image, $I_{\text{fake}}$. At this stage, an essential feedback loop is initiated, incorporating iterative adjustments informed by both perceptual and adversarial losses. The perceptual loss, $L_{\text{content}}$, is determined with the VGG19 model by comparing $I_{\text{fake}}$ with $I_{\text{att}}$. $G$ is adjusted to minimize this loss, thus improving the fidelity of the content. Simultaneously, the discriminator $D$ assesses the adversarial robustness by computing the adversarial loss, $L_{\text{adv}}$, with $I_{\text{fake}}$ and $I_{\text{att}}$. The aim of $D$ is to maximize $L_{\text{adv}}$, which provides essential feedback to $G$ for the ongoing enhancement of $I_{\text{fake}}$. This iterative method proceeds until a convergence condition is satisfied, resulting in the final image, $I_{\text{rec}}$, a polished version of $I_{\text{fake}}$.

\subsection{Residual Blocks}

Residual blocks play a vital role in deep neural network architectures, mainly by tackling vanishing gradient problems during training and enabling the learning of more intricate functions. These blocks are especially effective for tasks involving images, like image reconstruction and classification. The output $y$ of a residual block can be expressed using the following formula:

 \begin{equation}
     y = F(x, {W_i}) + x ,
 \end{equation}

where $x$ is the input vector to the residual block, $F(x, {W_i})$ represents the residual function applied to the input. This function typically involves one or more weighted layers within the block, including convolutional layers, activation functions, and normalization layers. ${W_i}$ denotes the weights associated with these layers. The operation $+$ is an element-wise addition, ensuring that the residual function and the input-output are of the same dimensions.

This research highlights the essential function of residual blocks in improving both the capabilities and the efficiency of the image reconstruction generator in a GAN structure.

Residual blocks facilitate the direct transmission of features from the network's early layers to its deeper layers. This type of feature propagation is highly advantageous in image reconstruction, where the detail in each layer plays a critical role.

Residual blocks facilitate the network in learning identity mappings, where the ideal output corresponds to an unchanged version of the input. In image reconstruction, if a segment of an image doesn't need alterations, the network can swiftly detect and preserve it, allowing attention to concentrate on sections requiring changes.

\subsection{Generator and Discriminator}

The methodology for image reconstruction used in this research utilizes the Pix2pix algorithm\cite{isola2017image} as a foundational technique, taking advantage of the capabilities of cGANs. The Pix2Pix framework is comprised of two primary elements: a generator and a discriminator. The generator, designed with a U-Net-style architecture, seeks to produce images that appear authentic, while the discriminator, functioning as a PatchGAN classifier, is able to distinguish between authentic and artificially generated images.

\subsubsection{\textbf{U-Net}} Ronneberger et al. \cite{ronneberger2015u} first presented the U-Net architecture, a sophisticated deep learning model mainly applied in the field of image segmentation. This architecture is particularly recognized for its unique design that includes an encoder for extracting features and a decoder for reconstructing images. The encoder is composed of eight blocks, each containing a convolutional layer, batch normalization, and \textit{LeakyReLU} activation. In contrast, the decoder is made up of seven blocks, each comprising a deconvolutional layer, batch normalization, and \textit{ReLU} activation, concluding with a final deconvolutional layer.

An important development in this study is the incorporation of seven residual blocks within each layer of the encoder's downsampling process. These residual blocks are purposefully designed to improve information flow using skip connections. Such connections are essential for mitigating the issue of gradient vanishing, which is often faced in deep neural networks, thus ensuring stable training and boosting the model's learning effectiveness. Each residual block effectively passes feature maps from one layer to the next, offering an extra layer of nonlinear processing to detect more intricate and delicate patterns in the images.

\subsubsection{\textbf{PatchGAN}} Introduced by Isola et al. in 2017\cite{isola2017image}, PatchGAN is a discriminator model within GANs. Unlike traditional discriminators that deliver a single scalar value representing the authenticity of an entire image, PatchGAN divides the image into multiple smaller patches. Each patch is assessed as real or fake, resulting in a two-dimensional array of probabilities for each patch. PatchGAN's advantage lies in its ability to provide more specific feedback to the generator about which regions of the image need enhancement. This targeted feedback helps the generator create more lifelike images with enhanced detail and texture.

The main function of the discriminator \textit{D} is to differentiate between the authenticity of a generated image \textit{G(x)}—marking it as false—and a real image, accurately identifying it as real. This evaluation occurs within the framework of the contour map \textit{x}, enabling \textit{D} to effectively judge the quality and realism of the generated image in relation to its original input.

This research improves the core design of the original PatchGAN discriminator \textit{D} by merging in residual blocks. In particular, each of the discriminator's convolutional layers is supplemented with seven residual blocks. This change notably boosts the efficiency of the feature extraction and discrimination processes of the discriminator. Through this approach, the discriminator's ability to analyze and accurately determine image features is significantly enhanced, allowing for more effective and precise distinction between authentic and synthetic images.

\subsection{Perceptual Feature Evaluation}

This study leverages high-level semantic attributes during image reconstruction by using a perceptual loss function to improve the quality of images, moving beyond dependence on basic pixel-level values. Perceptual loss, a deep learning loss function for tasks involving images and videos \cite{johnson2016perceptual}, evaluates differences between images or videos based on perceptual similarity instead of sole pixel discrepancies. The approach involves extracting features from various layers of both generated and target images, utilizing the pre-trained VGG19 model \cite{simonyan2014very} for this purpose. The perceptual loss is computed across all layers, except the topmost, ensuring the reconstructed images retain visual consistency and semantic fidelity.

\begin{equation}
    {\cal{L}}_{perc} (G) = \sum_k a_k || V_k (y) - V_k (G(x, z))||_1 ,
\end{equation}

where $V$ denotes the pre-trained VGG19 model, which is utilized to assess the perceptual quality of images, the variable $k$ represents the kth layer of the generated image $G(x, z)$ and the target image $y$, where the analysis of perceptual loss is conducted. The term $ak$ refers to taking the mean of the differences between the features extracted from the target image and those from the generated image at the corresponding layers, facilitating a layer-specific quantification of perceptual discrepancies.

\subsection{Overall Objective}
\noindent
\textbf{Generator Loss}. The generator's loss is generally described as the gap between the target label, usually set to 1 (indicating a real image), and the discriminator's prediction of the generator's output. The generator aims to reduce this loss by producing increasingly realistic samples that deceive the discriminator into believing they are genuine \cite{pan2020loss}.

In a conditional GAN, the generator's BCE (binary cross-entropy) loss can be formulated as follows:

\begin{equation}
    {\cal{L}}_{cGAN} (G) = -\frac{1}{N} \sum [log(D(G(x, z)))] ,
\end{equation}

The pixel loss (also known as L1 loss) measures the difference between the generated and target images. The equation for pixel loss is as follows:

\begin{equation}
    {\cal{L}}_{L1} (G) = ||y - G(x, z)||_1 ,
\end{equation}

where G(x, z) is the generated image from random noise x and a conditional variable z, y is the target image, and $||.||_1$ denotes the $L1$ norm.

In other words, pixel loss is the sum of the absolute differences between each pixel in the generated image and the corresponding pixel in the target image. The goal of training the Pix2Pix model is to minimize this pixel loss during the training process, encouraging the model to generate images as close as possible to the target images.

\noindent
\textbf{Discriminator Loss.} The discriminator loss of CGAN is typically defined as a binary cross-entropy loss, given by the following equation:

\begin{equation}
\begin{split}
    {\cal{L}}_{cGAN} (D) = -\frac{1}{N} \sum [(log(D(x, y))) + \\
    (log(1-D(G(x, z), y)))] ,
\end{split}
\end{equation}

where $N$ is the number of samples in the batch, $\sum$ denotes summation over all samples in the batch, $x$ is an input image, $z$ is a random noise vector or a conditional label, $y$ is a target output image, $G$ is the generator function that takes $x$ and $z$ as input and generates fake samples $G(x,z)$, $D$ is the discriminator function that takes real and fake samples and $z$ as input and outputs a probability score for each sample.

The initial term signifies the discriminator's loss in accurately identifying real data samples, while the subsequent term indicates the loss related to correctly classifying the generated data samples.

\noindent
\textbf{Objective function.} The objective function of our proposed method is to learn a mapping function from an input image to an output image using the pix2pix algorithm. The final objective function is a weighted sum of three losses:

\begin{equation}
\begin{split}
    G^* = \min_G \max_D {\cal{L}}_{cGAN} (G,D) + \lambda_1 {\cal{L}}_{L1} (G) \\
     + \lambda_2 {\cal{L}}_{perc} (G)
\end{split}
\end{equation}

Here, $\lambda$ is a hyperparameter that controls the trade-off between pixel loss and perceptual loss. By adjusting the value of $\lambda$, we can control the quality and the sharpness of generated images. In specific experiments, we found that the recovery of images is better when the value of $\lambda_1$ is 100 and the value of $\lambda_2$ is 1.

\section{Experiments}
\subsection{Experimental Setting}
\noindent
\textbf{Target models.} In our study, we select the InceptionV3 model \cite{szegedy2016rethinking}, the ResNet50V2 model \cite{he2016identity}, and the InceptionResNetV2 \cite{szegedy2017inception}, each pre-trained on the ImageNet dataset, as target models, and we perform adversarial attacks on them.

\noindent
\textbf{Attack models.} We applied six standard attacks—FGSM, BIM, PGD, C\&W, MI-FGSM, and AutoAttack—leveraging the attack function of Cleverhans\cite{papernot2018cleverhans}. The AutoAttack implementation specifically utilized Adversarial Robustness Toolbox v1.2.0 (ART)\cite{art2018}. Our results showed that enhancing the intensity of the attacks in the training data set improved the generative model's performance. Consequently, we increased $\epsilon$ to 16/255 for FGSM, BIM, PGD, MI-FGSM, and AutoAttack to increase the image production capabilities of the generative model. We applied a norm value of np.inf for an adversarial attack constrained by the $L\infty$-norm. Furthermore, we executed a total of 40 iterations (nb\_iter=40), each modifying the perturbation along the loss function's gradient at a step size of 0.01 (eps\_iter=0.01). For the C\&W attack, we utilized the $L2$-norm constraint as originally proposed in the literature.

\noindent
\textbf{Datasets.} We trained our proposed model and tested it on four popular image datasets, the Modified National Institute of Standards and Technology (MNIST)\cite{lecun1998mnist}, Fashion-MNIST \cite{xiao2017fashion}, CIFAR-10 dataset \cite{krizhevsky2009learning}, and the ImageNet Large Scale Visual Recognition Challenge (ILSVRC) 2012 dataset \cite{ILSVRC15}.

\begin{enumerate}
    \item \textbf{MNIST:} The (MNIST) dataset consists of 60,000 training and 10,000 test images, each of which is a grayscale image of 28x28 pixels. We trained the proposed model on the combined dataset (180,000 images) of all 60,000 training images under three attacks (FGSM, PGD, C\&W) and tested on all 10,000 testing images.
    \item \textbf{Fashion-MNIST:} Fashion-MNIST is an alternative to the traditional MNIST dataset for benchmarking machine learning algorithms. It contains 60,000 training and 10,000 testing grayscale images, each 28x28 pixels in size, representing ten different categories of fashion items such as shirts, sandals, and bags. In the same way as the MNIST dataset, we trained the proposed model on the combined dataset (180,000 images) of all 60,000 training images and tested it on all 10,000 testing images.
    \item \textbf{CIFAR-10:} The CIFAR-10 dataset comprises 60,000 32x32 color images spread across ten classes, with 6,000 images per class. The dataset is divided into 50,000 training images and 10,000 test images. We trained the proposed model on the combined dataset (150,000 images) of all 50,000 training images and tested it on all 10,000 testing images.
    \item \textbf{ImageNet ILSVRC2012:} The ImageNet ILSVRC2012 dataset contains more than one million high-resolution images distributed over 1,000 categories. We resized all images to 256*256 to comply with the input size of our model. We randomly selected 30 images from every 1000 categories in the ImageNet training set. Every set of 5 images underwent one adversarial attack (a total of 6 types of attacks), resulting in a training set encompassing 18,000 images. We randomly selected 5 images from each category of ILSVRC2012's testing set as our testing dataset (5,000 images).
\end{enumerate}

\noindent
\textbf{Comapred methods.} In this research, we assess the effectiveness of our newly proposed technique by comparing it with several established methods across various datasets, aiming for a thorough evaluation. For datasets with low resolution, such as MNIST, Fashion-MNIST (F-MNIST), and CIFAR-10, we contrast our method with Defense-GAN \cite{samangouei2018defense} and the Reconstruction method by Zhang \cite{zhang2021defense}. For the high-resolution dataset from ImageNet ILSVRC2012, we broaden our comparison to a wider array of techniques, specifically including Zhang's Reconstruction method, the Randomization method \cite{xie2017mitigating}, the Pixel Deflection method \cite{prakash2018deflecting}, the Super Resolution method \cite{mustafa2019image}, and the Wavelet Denoising method \cite{donoho1995noising}. This diverse range of comparisons aims to thoroughly test the robustness and efficacy of our method against adversarial attacks, showcasing its potential advantages in both low and high-resolution contexts.

\noindent
\textbf{Implemetation details.} The experiment for this study was implemented utilizing the NVIDIA RTX A6000 graphics processing unit. The programming language of choice was Python. The deep learning experiments were conducted using TensorFlow 2, selected for its comprehensive and adaptable ecosystem supporting cutting-edge machine learning developments.


\subsection{Main Results}
\subsubsection{\textbf{Performance on low-resolution datasets}}

\begin{table*}
    \centering
    \renewcommand\arraystretch{1.4}
    \tabcolsep=0.16cm
    \captionsetup{textformat=simple}
    \caption{Comparative evaluations for various methods on different datasets under FGSM, PGD ($\varepsilon=0.3$), and C\&W attacks. The abbreviations denote the following evaluation metrics: Acc refers to the accuracy (\%), PSNR refers to the Peak Signal-to-Noise Ratio (dB), Pr\_Time refers to the processing time (ms), and Tr\_Epochs refers to the training epochs. The best restoration performance for each attack is highlighted in \textbf{bold}.}
    \footnotesize
    \begin{tabular}{cccccccccccccc}
    \toprule
    \multirow{2}{*}{Dataset} & \multirow{2}{*}{Method} & \multicolumn{4}{c}{FGSM} & \multicolumn{4}{c}{PGD} & \multicolumn{4}{c}{C\&W} \\
    \cmidrule(r){3-6} \cmidrule(r){7-10} \cmidrule(r){11-14}
    && Acc & PSNR & Pr\_Time & Tr\_Epochs & Acc & PSNR & Pr\_Time & Tr\_Epochs & Acc & PSNR & Pr\_Time& Tr\_Epochs \\
    \toprule
    \multirow{5}{*}{MNIST} & Clean & 98.7 & - & - & - & 98.7 & - & - & - & 98.7 & - & - & -  \\
    & Attack & 27.4 & 30.31 & - & - & 9.3 & 30.87 & - & - & 12.6 & 29.39 & - & - \\
    & Defense-GAN & 96.6 & 34.48 & 2450 & 30,000 & 95.8 & 34.16 & 2450 & 30,000 & 96.5 & \textbf{34.93} & 2450 & 30,000 \\
    & Reconstruction & 96.4 & \textbf{34.73} & 6.7 & \textbf{1000} & 95.4 & 34.65 & 6.8 & \textbf{1000} & 94.0 & 34.42 & 6.7 & \textbf{1000} \\
    & Ours & \textbf{98.6} & 34.27 & \textbf{2.1} & \textbf{1000} & \textbf{97.8} & \textbf{34.87} & \textbf{2.2} & \textbf{1000} & \textbf{98.0} & 34.02 & \textbf{2.2} & \textbf{1000} \\
    \hline
    \multirow{5}{*}{F-MNIST} & Clean & 88.8 & - & - & - & 88.8 & - & - & - & 88.8 & - & - & -  \\
    & Attack & 10.3 & 29.98 & - & - & 2.0 & 29.18 & - & - & 1.6 & 27.38 & - & - \\
    & Defense-GAN & 87.5 & 30.69 & 2630 & 30,000 & 82.3 & 30.53 & 2635 & 30,000 & 74.0 & 30.25 & 2640 & 30,000 \\
    & Reconstruction & 87.1 & 31.93 & 6.6 & \textbf{1000} & 76.3 & \textbf{31.85} & 6.5 & \textbf{1000} & \textbf{84.7} & \textbf{31.57} & 6.8 & \textbf{1000} \\
    & Ours & \textbf{88.3} & \textbf{31.98} & \textbf{2.1} & \textbf{1000} & \textbf{87.3} & 31.53 & \textbf{2.1} & \textbf{1000} & 79.1 & 31.24 & \textbf{2.2} & \textbf{1000} \\
    \hline
    \multirow{5}{*}{CIFAR-10} & Clean & 89.6 & - & - & - & 89.6 & - & - & - & 89.6 & - & - & -  \\
    & Attack & 8.3 & 27. 43 & - & - & 1.3 & 27.89 & - & - & 0.2 & 27.46 & - & - \\
    & Defense-GAN & 9.6 & 29.46 & 3750 & 50,000 & 9.5 & 29.82 & 3750 & 50,000 & 8.5 & 29.57 & 3750 & 50,000 \\
    & Reconstruction & 90.3 & 34.07 & 28 & \textbf{350} & 88.9 & \textbf{33.97} & 27 & \textbf{350} & 87.5 & \textbf{33.58} & 28 & \textbf{400} \\
    & Ours & \textbf{94.6} & \textbf{34.14} & \textbf{3.5} & 400 & \textbf{92.9} & 33.82 & \textbf{3.5} & 400 & \textbf{91.6} & 33.37 & \textbf{3.5} & \textbf{400} \\
    \bottomrule
    \end{tabular}
\end{table*}

Table I offers a detailed evaluation of various defense strategies against FGSM, PGD, and C\&W attacks, focusing on datasets such as MNIST, F-MNIST, and CIFAR-10. The metrics used for evaluation include Accuracy (Acc), Peak Signal-to-Noise Ratio (PSNR), Processing Time (Pr\_Time), and Training Epochs (Tr\_Epochs). 

For the MNIST dataset under FGSM attack, our approach achieves an impressive accuracy of 98.6\% with a minimal processing time of 2.1 ms. It only needs 1000 training epochs, marking a significant enhancement over other methods. Moreover, for PGD and C\&W attacks, our approach consistently outperforms with accuracies of 97.8\% and 98.0\% respectively, coupled with the lowest processing times and consistent training demands.

Regarding the F-MNIST dataset, our model also achieves high performance, maintaining an accuracy of 88.3\% when subjected to FGSM, increasing to 87.3\% with PGD, and showing a small decrease to 79.6\% under the C\&W attack. It balances swift processing with minimal training epochs, demonstrating impressive efficiency and rapid adaptability.

Our approach exhibits substantial robustness on the CIFAR-10 dataset, achieving peak accuracies of 94.6\% when tested with FGSM, 92.9\% with PGD, and 91.6\% with C\&W. Furthermore, it features the quickest processing times for all evaluations and achieves swift model training, requiring just 400 epochs for reliable convergence.

\subsubsection{\textbf{Performance on ImageNet dataset}} 

\begin{table*}
\renewcommand\arraystretch{1.5}
\footnotesize
\caption{The classification accuracy (\%) of our method compared with state-of-the-art defense methods on ImageNet dataset. The abbreviations denote the defensive methods: 'Random' refers to the Randomization Method, 'PD' represents the Pixel Deflection Method, 'WD' corresponds to the Wavelet Denoising Method, and 'SR' signifies the Super-Resolution Method. They are evaluated against seven adversarial attacks. The best restoration performance for each attack is highlighted in \textbf{bold}, and the second is with \underline{underline}.}\label{tab2}
\begin{center}
\tabcolsep=0.2 cm
\begin{tabular}{ccc|cccccccc}
\toprule
Attack & Model & No Defense & Random \cite{xie2017mitigating} & PD \cite{prakash2018deflecting} & SR \cite{mustafa2019image} & WD \cite{donoho1995noising} & WD+PD & WD+SR & Recons \cite{zhang2021defense} & Ours \\
\toprule
\multirow{3}{*}{Clean} &  InceptionV3 &  76.8 & \underline{74.7} & 67.1 & 73.8 & 72.1 & 70.5 & \underline{74.7} & 72.8 & \textbf{76.1} \\
&  Resnet50V2 &  67.0 & 61.9 & 59.4 & 62.4 & 62.1 &  62.0 & \textbf{62.8} & \underline{62.6} & 62.4 \\
&  InceptionResnetV2 &  79.3 & 76.2 & 69.6 & 77.1 & 74.5 & 73.0 & \textbf{77.8} & 75.2 & \underline{76.5} \\
\hline
\multirow{3}{*}{FGSM} &  InceptionV3 &  25.8 & 44.8 & 27.6 & 62.8 & 27.9 & 28.3 & \underline{64.5} & \textbf{64.9} & 63.8 \\
&  Resnet50V2 & 17.8 & 37.3 & 17.0 & 46.5 & 8.5 & 17.8 & \underline{48.6} & \textbf{50.1} & 45.9 \\
&  InceptionResnetV2 & 21.0 & 46.8 & 28.1 & 64.9 & 45.6 & 44.1 & 66.3 & \underline{68.5} & \textbf{70.1} \\
\hline
\multirow{3}{*}{BIM} &  InceptionV3 &  1.8 & 67.9 & 12.9 & 67.5 & 61.7 & 59.3 & \underline{68.9} & 67.7 & \textbf{70.3} \\
&  Resnet50V2 & 0.0 & 51.1 & 9.7 & 52.1 & 49.1 & 50.5 & \textbf{55.3} & 54.1 & \underline{54.3} \\
&  InceptionResnetV2 & 7.8 & 68.1 & 13.4 & 69.3 & 62.4 & 61.7 & 71.6 & \underline{71.9} & \textbf{73.8} \\
\hline
\multirow{3}{*}{PGD} &  InceptionV3 &  2.4 & \underline{67.3} & 12.4 & 65.0 & 59.7 & 59.4 & 67.1 & 66.5 & \textbf{69.8} \\
&  Resnet50V2 & 1.3 & 52.7 & 9.6 & 53.5 & 29.1 & 27.5 & 53.1 & \textbf{56.8} & \underline{53.8} \\
&  InceptionResnetV2 & 8.1 & 69.7 & 12.7 & 68.7 & 60.9 & 60.2 & 71.7 & \underline{72.8} & \textbf{73.7} \\
\hline
\multirow{3}{*}{MI-FGSM} &  InceptionV3 &  0.1 & 68.3 & 11.1 & 70.1 & 59.1 & 60.1 & \textbf{70.6} & 63.8 & \underline{66.7} \\
&  Resnet50V2 & 0.0 & \textbf{58.9} & 10.1 & 56.9 & 47.6 & 48.3 & 57.6 & 50.2 & \underline{58.4} \\
&  InceptionResnetV2 & 8.5 & 70.0 & 13.0 & 69.7 & 63.9 & 61.7 & \textbf{72.3} & 66.8 & \underline{70.2} \\
\hline
\multirow{3}{*}{C\&W} &  InceptionV3 &  0.0 & 68.7 & 3.9 & 70.4 & 58.5 & 57.4 & \textbf{73.7} & 66.1 & \underline{71.6} \\
&  Resnet50V2 & 0.0 & 58.6 & 1.7 & 58.6 & 52.9 & 53.4 & \textbf{62.3} & 60.9 & \underline{62.1} \\
&  InceptionResnetV2 & 4.4 & 69.8 & 4.7 & 71.5 & 64.2 & 61.5 & \textbf{75.8} & 69.4 & \underline{72.1} \\
\hline
\multirow{3}{*}{AutoAttack} &  InceptionV3 &  0.5 & 63.7 & 7.3 & 64.7 & 58.1 &  57.3 & \underline{67.4} & 65.3 & \textbf{70.4} \\
&  Resnet50V2 & 0.0 & 51.4 & 13.8 & 56.9 & 47.7 & 47.9 & \underline{57.9} & 56.8 & \textbf{61.5} \\
&  InceptionResnetV2 & 0.4 & 64.0 & 12.6 & 67.1 & 59.0 & 60.9 & \underline{69.3} & 68.0 & \textbf{71.2} \\
\hline
\multirow{3}{*}{Deepfool} &  InceptionV3 & 0.4 & 68.3 & 11.0 & 70.3 & 57.5 & 57.6 & 70.8 & \textbf{74.1} & \underline{72.3} \\
&  Resnet50V2 & 0.0 & 60.9 & 9.9 & 59.8 & 48.1 & 50.1 & 57.7 & \underline{64.2} & \textbf{64.9} \\
&  InceptionResnetV2 & 0.7 & 71.1 & 13.1 & 72.6 & 62.7 & 64.3 & 73.1 & \textbf{77.7} & \underline{75.8} \\

\bottomrule
\end{tabular}
\end{center}
\end{table*}

Table II presents a comparative evaluation of the classification accuracy (\%) for different defense strategies against several adversarial attacks on the ImageNet dataset. The defenses examined encompass Randomization (Random), Pixel Deflection (PD), Super-Resolution (SR), Wavelet Denoising (WD), and combinations of WD with PD (WD+PD) and SR (WD+SR), alongside our proposed technique. This table assesses the effectiveness of these methods in both clean conditions and under adversarial attacks, such as FGSM, BIM, PGD, MI-FGSM, C\&W, AutoAttack, and Deepfool.

When evaluated in clean scenarios, our technique reliably delivers exceptional accuracy for all three architectures, achieving 76.1\%, 62.4\%, and 76.5\% using InceptionV3, Resnet50V2, and InceptionResnetV2, respectively. This demonstrates its robustness against adversarial challenges while maintaining high accuracy with unaltered input, emphasizing the method's practicality without sacrificing standard performance.

In the context of the FGSM attack, our approach significantly exceeds other methods, with the InceptionResnetV2 model attaining a top accuracy of 70.1\%. The Recons \cite{zhang2021defense} approach closely trails, exhibiting notable performance, especially with InceptionResnetV2 at 68.5\%. These findings highlight the efficacy of combining denoising and super-resolution techniques to counteract simple, quick-gradient attacks.

Our approach continues to outperform in more advanced iterative attacks such as BIM and PGD, especially with InceptionResnetV2, where it achieves accuracies of 73.8\% and 73.7\%, respectively. This marks a substantial enhancement over the baseline methods, which find it challenging to surpass the 70\% mark in such circumstances. 

Our approach notably outperforms among diversified attack strategies such as MI-FGSM and C\&W, particularly when applied to the InceptionResnetV2 model, achieving accuracies of 70. 2\% and 72. 1\%, respectively. This shows the proficiency of the method in overcoming gradient masking and attacks utilizing mixed approaches to deceive defensive models.

Finally, when tested against AutoAttack and Deepfool, which specialize in overcoming standard defense strategies by utilizing adaptive and deep exploration techniques, our approach maintains strong performance. With the InceptionResnetV2 model, it achieves accuracy rates of 71.2\% and 75.8\%, respectively, surpassing other methods significantly. This confirms its resilience against the most formidable automated adversarial attacks.

This comprehensive assessment underscores the advantages of our proposed approach in multiple adversarial scenarios, demonstrating its effectiveness in enhancing model robustness against a wide range of attacks while preserving impressive accuracy in non-adversarial environments. The findings indicate that our method could serve as a reliable standard for crafting future defense strategies in adversarial machine learning.

\subsubsection{\textbf{Model generalizability}}

\begin{table*}
\renewcommand\arraystretch{1.7}
\tabcolsep=0.22cm
\caption{The classification accuracy (Acc, \%), and generalizability (G) of six pre-trained (on a single adversarial dataset) image reconstruction models, namely ${Model}_{FGSM}$, ${Model}_{BIM}$, ${Model}_{PGD}$, ${Model}_{MI}$, ${Model}_{C\&W}$, and ${Model}_{AA}$. Those models are evaluated against seven types of adversarial attacks. The best performance of accuracy is highlighted in \textbf{bold}.}\label{tab1}
\begin{center}
\footnotesize
\begin{tabular}{cc|cccccccccccc}
\toprule
\multirow{2}{*}{Attacks($\epsilon=0.01$)} & \multirow{2}{*}{No defense} & \multicolumn{2}{c}{${Model}_{FGSM}$} & \multicolumn{2}{c}{${Model}_{BIM}$} & \multicolumn{2}{c}{${Model}_{PGD}$} & \multicolumn{2}{c}{${Model}_{MI}$} & \multicolumn{2}{c}{${Model}_{C\&W}$} & \multicolumn{2}{c}{${Model}_{AA}$} \\
&& Acc $\uparrow$ & G $\downarrow$ & Acc $\uparrow$ & G $\downarrow$ & Acc $\uparrow$ & G $\downarrow$ & Acc $\uparrow$ & G $\downarrow$ & Acc $\uparrow$ & G $\downarrow$ & Acc $\uparrow$ & G $\downarrow$ \\
\toprule
Clean & 76.8 & 74.8	& - & \textbf{76.5} & - & 76.4 & - & 75.6 & - & 75.7 & - & 73.5 & -  \\
FGSM & 25.8 & \textbf{69.2} & - & 63.5	& 0.116 & 68.3 & 0.039 & 69.1 & 0.034 & 65.6 & 0.064 & 64.5 & 0.089 \\
BIM & 1.8 & 72.4 & -0.046 & 71.8 & - & \textbf{73.4} & -0.032 & 72.4 & -0.013 & 71.1 & -0.014 & 69.1 & 0.024 \\
PGD & 2.4 & 71.4 & -0.032 & 70.8 & 0.014 & 71.1 & - & \textbf{72.6} & -0.015 & 71.2 & -0.016 & 68.7 & 0.030\\
MI-FGSM & 0.1 & 71.1 & -0.027 & 66.8 & 0.070 & 71.2 & -0.001 & \textbf{71.5} & - & 67.9 & 0.031 & 66.3 & 0.064 \\
C\&W & 0 & 67.4 & 0.026 & 68.6 & 0.045 & 66.9 & 0.060 & 68.0 & 0.049 & \textbf{70.1} & - & 66.0 & 0.068 \\
AutoAttack & 0.5 & 54.6 & 0.211 & 41.2 & 0.426 & 27.8 & 0.609 & 53.5 & 0.252 & 61.9 & 0.117 & \textbf{70.8} & -  \\
DeepFool & 0.4 & 70.3 & -0.016 & 72.1 & -0.004 & 73.8 & -0.038 & \textbf{73.9} & -0.034 & 72.7 & -0.037 & 72.1 & -0.018 \\
\bottomrule
\end{tabular}
\end{center}
\end{table*}

\begin{table}
\renewcommand\arraystretch{1.7}
\tabcolsep=0.3cm
\caption{The classification accuracy (Acc, \%)of our pre-trained model (trained on InceptionV3) in defending against adversarial attacks, targeting ResNet50 and InceptionResNetV2 models to assess the cross-model generalizability. }
    \centering
    \footnotesize
    \begin{tabular}{c|cc|cc}
    \toprule
        \multirow{2}{*}{Attack} & \multicolumn{2}{c|}{Resnet50V2} & \multicolumn{2}{c}{InceptionResnetV2} \\
         & No defense & Defense & No defense & Defense \\
    \toprule
       Clean  & 66.9 & 63.8 & 80.8 & 78.2 \\
        FGSM & 18.3 & 44.6 & 21.3 & 69.2 \\
        BIM & 1.1 & 48.1 & 8.4 & 73.8 \\
        PGD & 1.7 & 47.8 & 8.7 & 70.1 \\
        MI-FGSM & 0.0 & 47.2 & 8.9 & 75.5 \\
        C\&W & 0.0 & 51.6 & 5.1 & 69.5 \\
        AutoAttack & 0.0 & 40.1 & 0.4 & 74.1 \\
        DeepFool & 0.0 & 53.7 & 1.1 & 73.7\\
    \bottomrule
    \end{tabular}
    \label{tab:my_label}
\end{table}

Our research focused on assessing how well models maintain effectiveness when facing various adversarial scenarios. In machine learning, generalizability is defined as a model's ability to use the knowledge gained during training to handle new, unseen data situations effectively. This characteristic is crucial as it influences a model's resilience and flexibility with respect to shifts in data distributions frequently seen in practical applications. A model with strong generalizability performs well on both the training set and new, unfamiliar data. This is due to its ability to grasp the fundamental structures and patterns of the dataset, as opposed to just memorizing specific examples from the training phase. Consequently, such models are more likely to sustain good performance when confronted with fresh, unseen adversarial challenges.

\noindent \textbf{On different datasets.} We performed experiments employing six pre-trained image reconstruction models: ${Model}_{FGSM}$, ${Model}_{BIM}$, ${Model}_{PGD}$, ${Model}_{MI}$, ${Model}_{C\&W}$, and ${Model}_{AA}$. These models were tested on image restoration tasks in the face of seven adversarial attacks, among which DeepFool was regarded an unknown attack, having been absent from the training set. The aim was to determine whether these models could maintain their efficacy in countering adversarial attacks distinct from those they were trained on.

As illustrated in Table III, all models exhibited an enhancement in classification accuracy when facing adversarial scenarios that were not specifically anticipated during their training. This indicates that the image reconstruction models are not only adept at countering known adversarial attacks but also display a notable resilience to new adversarial tactics, which highlights their capacity for generalization.

\noindent \textbf{On different target models.} TABLE IV illustrates the classification accuracy of a defense mechanism applied to a model using InceptionV3, as evaluated against various adversarial attacks on two other models: ResNet50V2 and InceptionResNetV2. This study assesses the model's capacity to generalize across different architectures. Implementing defense strategies notably elevates accuracy rates for all kinds of attacks. For instance, accuracy against the FGSM attack is raised to 44.6\% for ResNet50V2 and 69.2\% for InceptionResNetV2. Some attacks, like Carlini \& Wagner (C\&W) and DeepFool, are more challenging defenses, yet accuracy still improves significantly from 0\% to 51.6\% and 53.7\% for ResNet50V2, respectively. These findings highlight the effectiveness of the defense mechanism, showcasing its ability to strengthen resistance against a broad spectrum of sophisticated adversarial attacks, thereby enhancing the robustness of the model and ensuring better adaptability across multiple architectures.

If training a model against various attack vectors improves its generalization, it implies that developing models capable of handling a wide array of attacks could lead to more adaptable and resilient defenses. This strategy would be particularly advantageous in real-world scenarios where unexpected adversarial challenges arise, enhancing the models' applicability and effectiveness. This aligns with the need for flexible security measures in machine learning applications.

\subsubsection{\textbf{Quantitative Evaluation}}

\begin{figure}[!ht]
    \centering
    \includegraphics[width=0.6\linewidth]{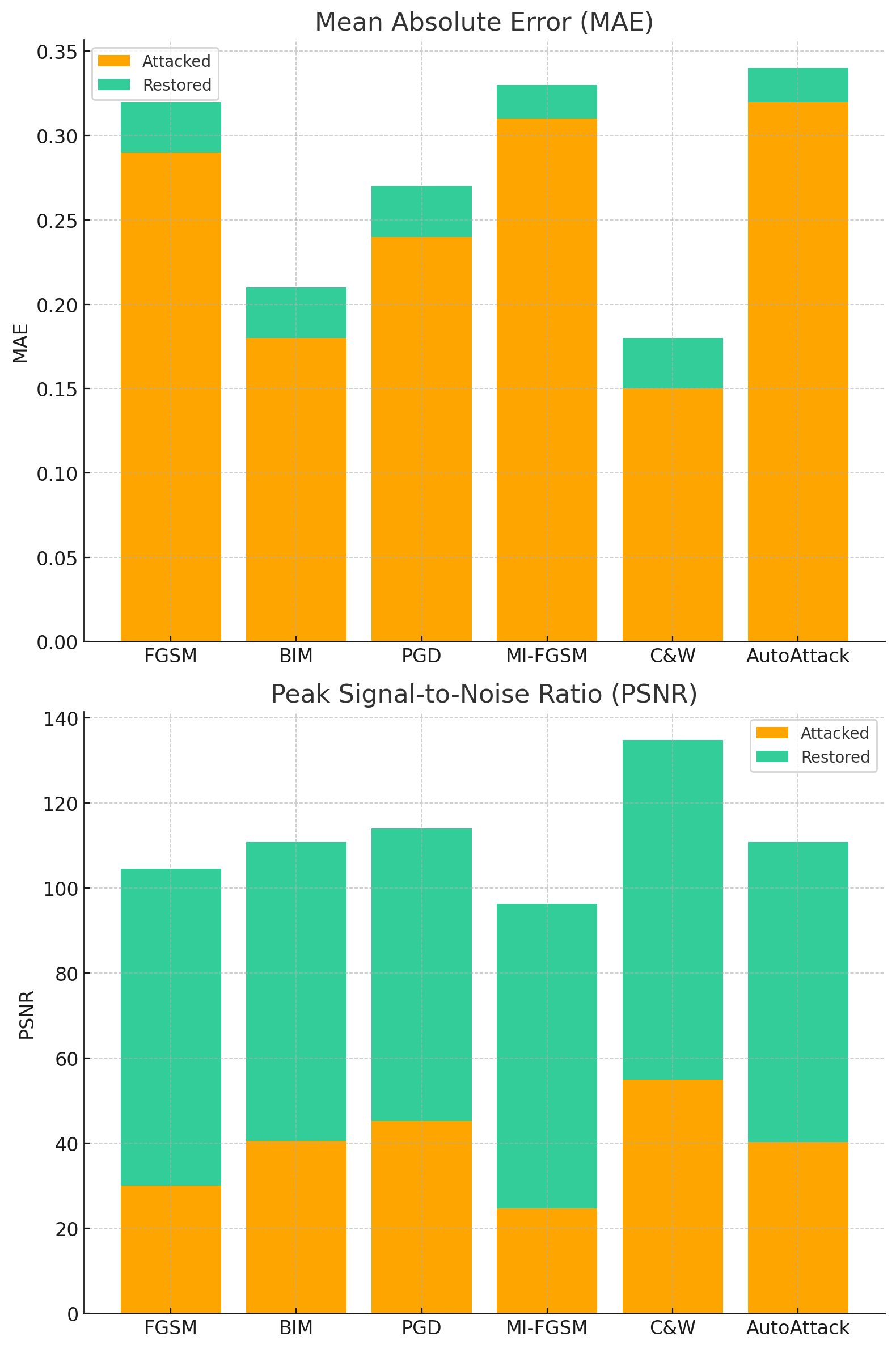}
    \caption{The computation of the PSNR and MAE values for both the images subjected to six types of adversarial attacks and those reconstructed by the universal defense model when compared to the original images.}
\end{figure}

In this study, we assess the effectiveness of our proposed defense model using two well-established quantitative evaluation metrics: Mean Absolute Error (MAE) and Peak Signal-to-Noise Ratio (PSNR). These metrics provide a robust measure of model performance in terms of both error reduction and image quality improvement following adversarial attacks.

As illustrated in Fig. 4, the MAE values indicate the average absolute error between the original and attacked or restored images across various adversarial attacks. The lower MAE values for restored images compared to attacked images across all attack methods showed significant error reduction due to our model's restoration capabilities. While attacks like C\&W and AutoAttack show the highest error increase upon attack, they also display substantial error reduction after restoration, highlighting our model's effectiveness against more sophisticated attacks.

The PSNR values, which measure the ratio of the maximum possible power of a signal to the power of corrupting noise, further affirm the model’s efficiency. Higher PSNR values for restored images than those attacked validate the model's ability to maintain image fidelity by effectively mitigating the impact of noise introduced by the adversarial attacks. The PSNR improvement is particularly notable in methods like FGSM and PGD, where the restoration process yields a near-complete recovery of signal integrity.

\subsubsection{\textbf{Robustness Check}}

\begin{figure}[!ht]
\centering
\subfloat{\includegraphics[width=0.23\textwidth]{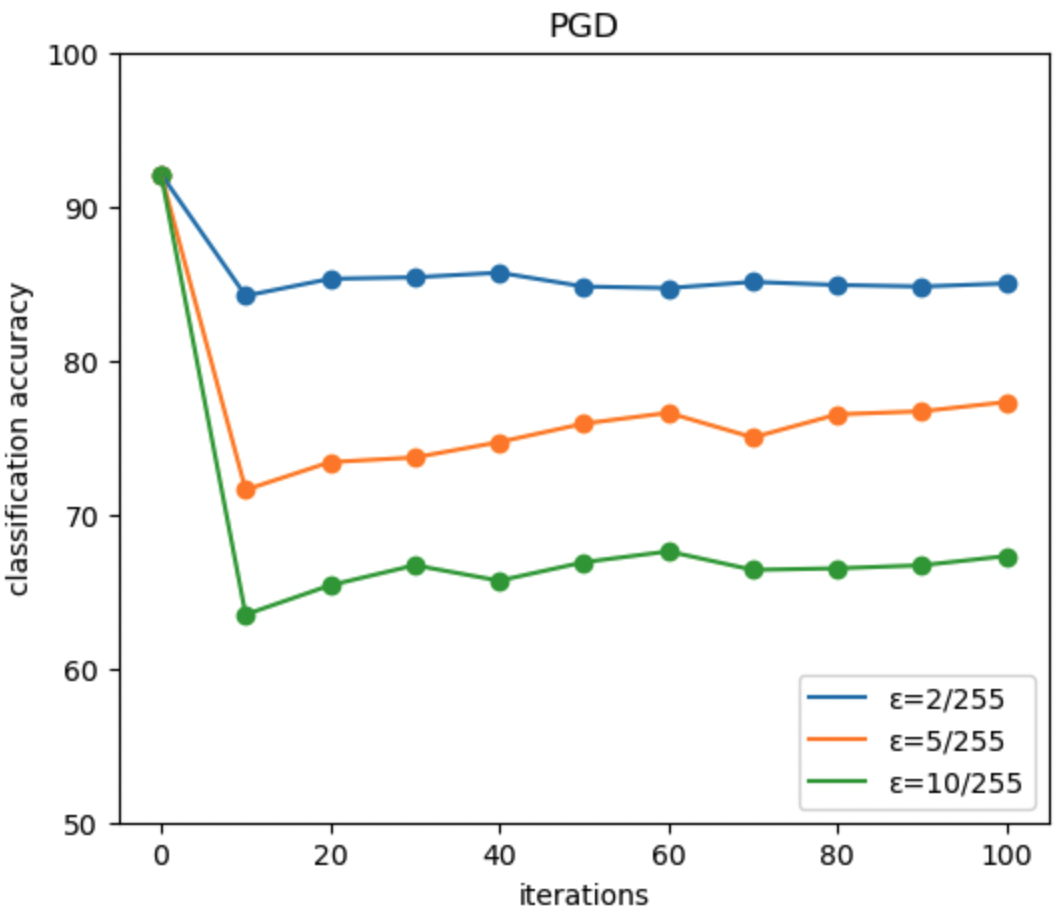}}
\hfill
\subfloat{\includegraphics[width=0.23\textwidth]{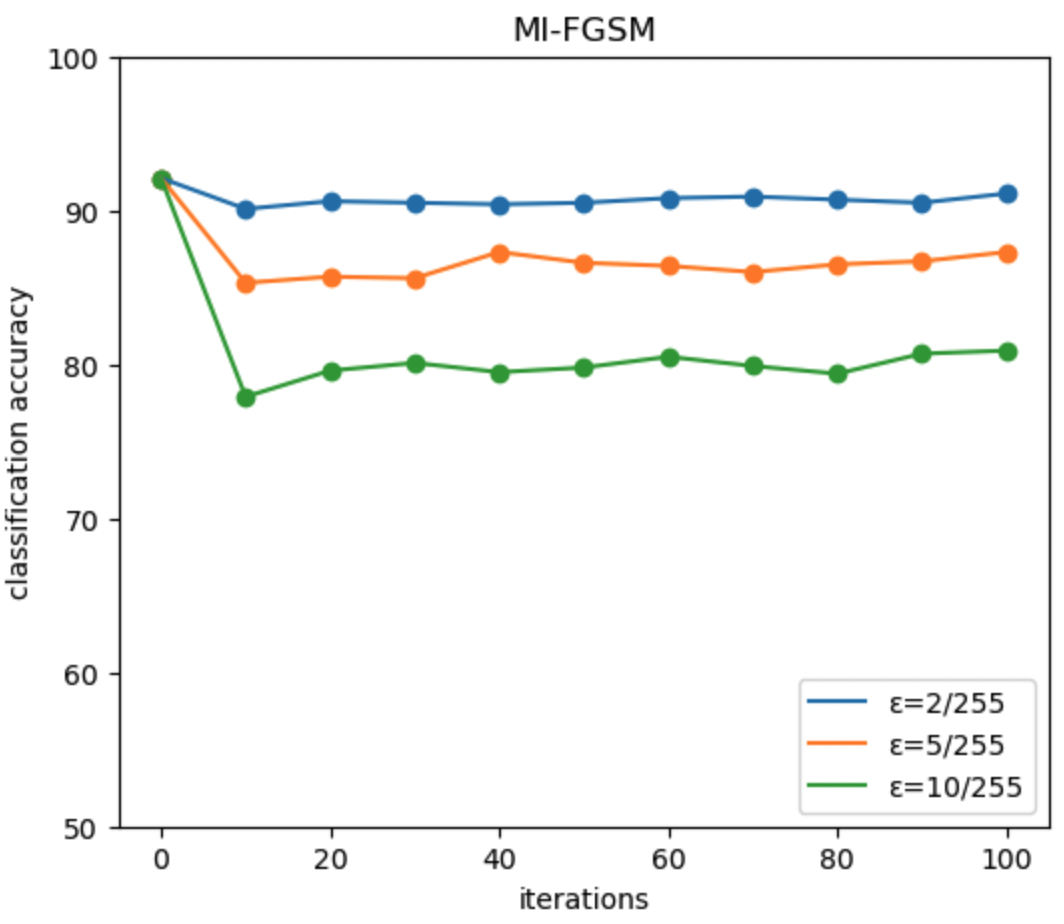}}
\caption{Robustness Check using the PGD attack and the MI-FGSM attack. To simulate different attack strengths, we gradually change the iteration number from 10 to 100, and the $\epsilon$ includes 2/255, 5/255, and 10/255.}
\end{figure}

For the robustness check, we thoroughly analyzed the behavior of our model against the PGD and MI-FGSM attacks by methodically varying both the iteration number (ranging from 10 to 100) and the perturbation strength, represented by $\epsilon$ values of 2/255, 5/255, and 10/255.

Fig. 5 shows that the model's classification accuracy rate remains relatively stable across different numbers of iterations and varying intensities of adversarial perturbations. This stability in performance underlines the resilience of our defense mechanism, suggesting that the model is robust and reliable in maintaining high classification accuracy under continuous and varying adversarial pressures.

Furthermore, these findings highlight the model's ability to effectively manage and mitigate the distortions caused by different attack strategies and intensities, ensuring that the integrity and reliability of classification remain intact even under advanced adversarial conditions. This level of sustained performance across various adversarial settings emphasizes the sophisticated defensive capabilities embedded within our model, positioning it as a formidable tool against diverse and dynamic adversarial threats in machine learning applications.

\subsection{Ablation Studies}

\begin{figure}[!ht]
    \centering
    \includegraphics[width=\linewidth]{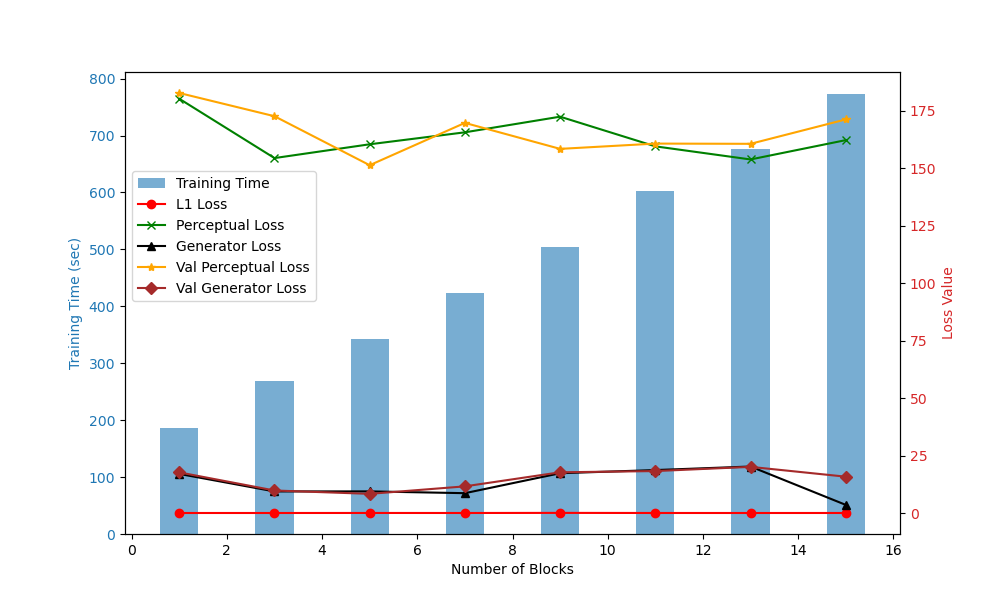}
    \caption{Impact of Residual Block Count on Model Performance and Training Efficiency. }
\end{figure}

\begin{figure*}
  \centering
  \subfloat[Training L1 loss]{
    \includegraphics[width=0.32\textwidth]{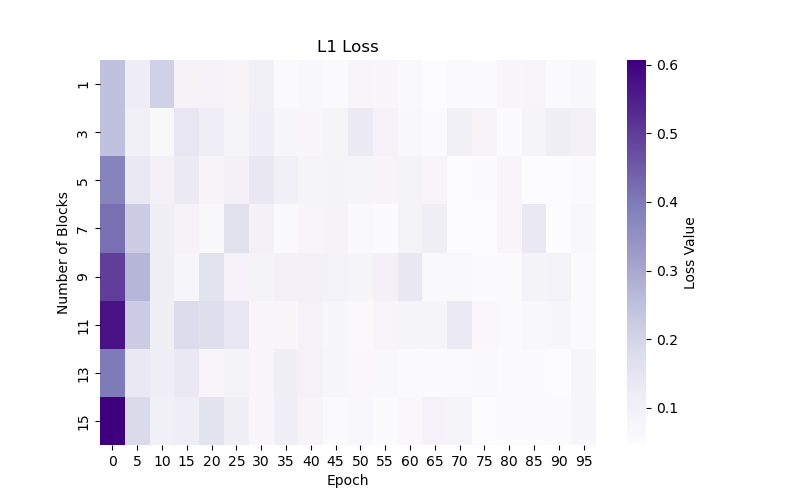}}
  \hfill
  \subfloat[Training Perceptual loss]{
    \includegraphics[width=0.32\textwidth]{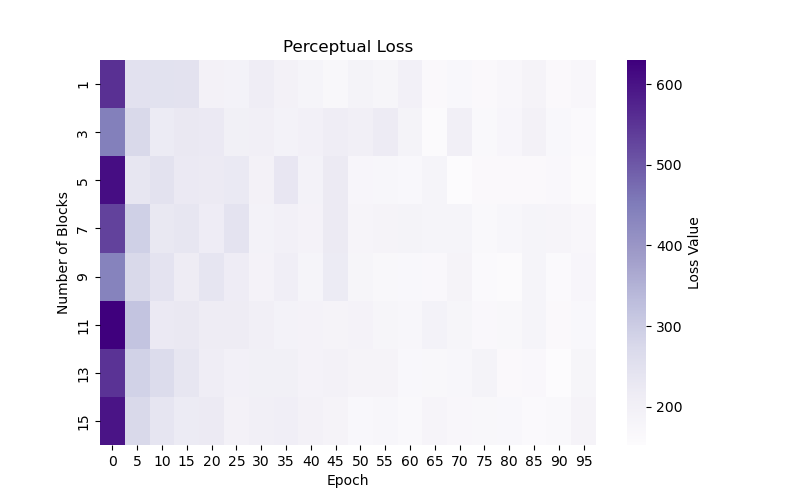}}
  \hfill
  \subfloat[Training Generator loss]{
    \includegraphics[width=0.32\textwidth]{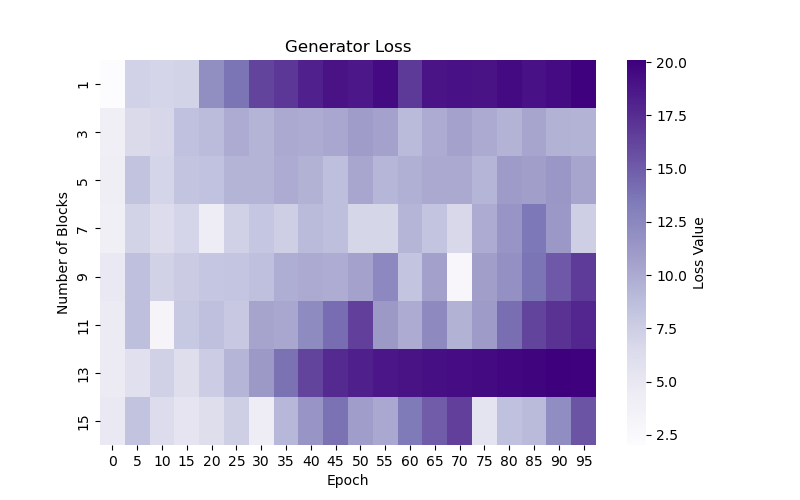}}

  \subfloat[Validation L1 loss]{
    \includegraphics[width=0.32\textwidth]{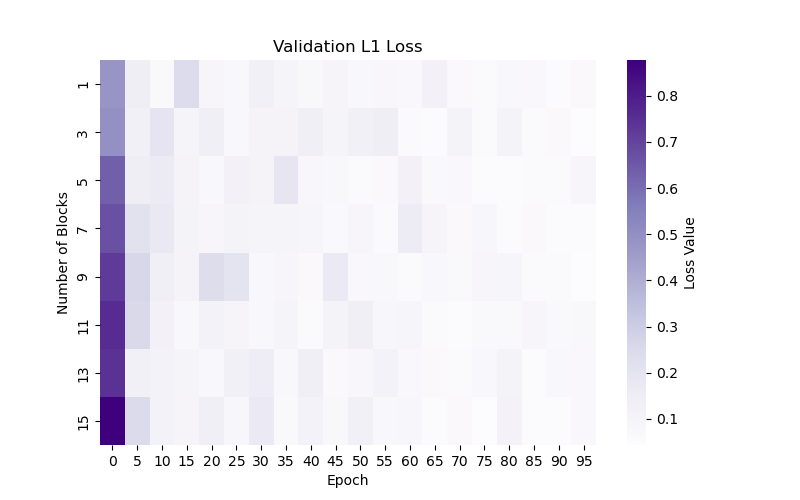}}
  \hfill
  \subfloat[Validation Perceptual loss]{
    \includegraphics[width=0.32\textwidth]{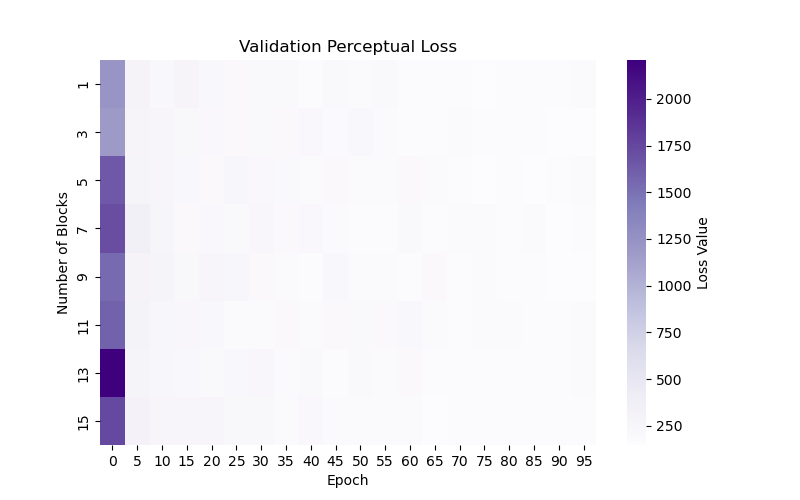}}
  \hfill
  \subfloat[Validation Generator loss]{
    \includegraphics[width=0.32\textwidth]{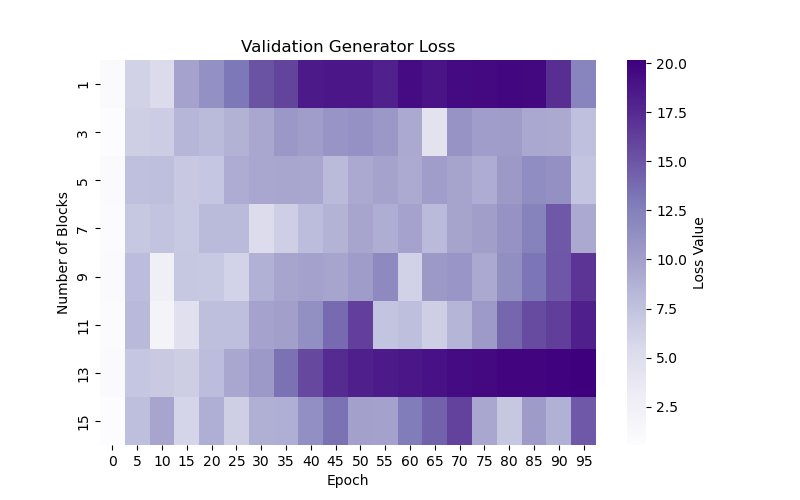}}   
  \caption{Loss functions according to different residual blocks under PGD attack.}
\end{figure*}

In our ablation study, depicted in Fig. 7, we thoroughly examined the optimal count of residual blocks required for maintaining consistent performance under adversarial scenarios. We analyzed how different numbers of residual blocks (1, 3, 5, 7, 9, 11, 13, 15) influence L1 loss, perceptual loss, and generator loss throughout training and validation. Our results show that though losses decline as the number of residual blocks increases, the improvements taper off beyond 7 blocks, indicating limited gains from additional complexity.

Moreover, our evaluation of training duration and computational effectiveness, as illustrated in Fig. 6, supports these results. We observed a notable increase in training time once the number of blocks exceeded 6, without any associated gains in loss metrics, highlighting the drawbacks of unnecessarily increasing the model's complexity.

Consequently, we determined that employing 7 residual blocks represents an optimal balance between model complexity, efficiency, and performance. This configuration minimizes the reconstruction error and perceptual loss and maintains reasonable training durations. The model consistently delivered high classification accuracy and superior image quality, which is imperative for practical deployments where performance and computational efficiency are paramount. This careful balancing act ensures the model's robustness against adversarial threats while supporting its feasibility for real-time application scenarios.

\begin{table}
    \centering
    \renewcommand\arraystretch{1.6}
    \tabcolsep=0.2cm
    \captionsetup{textformat=simple}
    \caption{The ablation study's image restoration accuracy (\%) on with/without residual blocks (number of 7).}
    \footnotesize
    \begin{tabular}{ccccc}
    \toprule
    Datasets & Attacks & No Defense & without & with \\
    \toprule
    \multirow{3}{*}{\makecell{MNIST \\ (98.7)}} & FGSM & 27.4 & 94.3 & \textbf{98.6} \\
    & PGD & 9.3 & 93.2 & \textbf{97.8} \\
    & C\&W & 12.6 & 94.5 & \textbf{98.0} \\
    \hline
    \multirow{3}{*}{\makecell{F-MNIST \\ (88.8)}} & FGSM & 10.3 & 84.8 & \textbf{88.3} \\
    & PGD & 2.0 & 83.9 & \textbf{87.3} \\
    & C\&W & 1.6 & 77.5 & \textbf{79.0} \\
    \hline
    \multirow{3}{*}{\makecell{CIFAR-10 \\ (89.6)}} & FGSM & 8.3 & 80.5 & \textbf{94.6} \\
    & PGD & 1.3 & 79.8 & \textbf{92.9} \\
    & C\&W & 0.2 & 78.0 & \textbf{91.6} \\
    \hline
    \multirow{3}{*}{\makecell{ImageNet \\ (76.8)}} & FGSM & 25.8 & 55.5 & \textbf{63.8} \\
    & PGD & 2.4 & 58.9 & \textbf{69.8} \\
    & C\&W & 0.0 & 61.7 & \textbf{71.6} \\
    \bottomrule
    \end{tabular}
\end{table}

Table V evaluates model performance under three different scenarios: without any defense, with a defense lacking residual blocks, and with a defense that includes residual blocks. 

For the MNIST and F-MNIST datasets, the result demonstrates a consistent trend wherein incorporating residual blocks substantially elevates defense efficacy. Specifically, under the FGSM attack, accuracy on MNIST escalates from 94.3\% to 98.6\% with the adoption of residual blocks. Likewise, in Fashion-MNIST, even under the more sophisticated C\&W attack, accuracy improves from 77.5\% to 79.0\% with the inclusion of residual blocks. 

Regarding CIFAR-10, the most significant enhancement is observed under the FGSM attack, where accuracy increases from 80.5\% to 94.6\% with the introduction of residual blocks. In the case of ImageNet, despite its complex scale and diversity, the defense incorporating residual blocks still significantly boosts recognition accuracy compared to configurations without residual blocks. For example, under the PGD attack, the inclusion of residual blocks yields an accuracy of 69.8\%, an improvement from 58.9\%.

\section{Futher Analysis}
Our method consistently shows exceptional or comparable results in all measures and datasets. It is characterized by high accuracy, notable PSNR values, rapid processing, and reliable training epochs, highlighting its robustness and effectiveness, particularly against adversarial FGSM attacks. However, a more thorough examination of the particular architecture variations, optimization strategies, and potential trade-offs among the models could yield a more complete understanding of their advantages and limitations.

\noindent
\textbf{Trade-off between Model Complexity and Efficiency.} Despite achieving the highest accuracy across various datasets, our method stands out by maintaining remarkable efficiency in processing times. This balance is both sought after and challenging to achieve, as higher accuracy typically involves increased complexity, resulting in longer processing durations. The efficiency of "Our Model" may be credited to a carefully optimized architecture or the implementation of successful pruning strategies.

\noindent
\textbf{Training Efficiency.} The number of epochs required for a model to achieve convergence often reflects its training efficiency. Our approach reliably converges in approximately 1,000 epochs across three different datasets, suggesting stable and efficient training performance. This efficiency can be attributed to improved optimization methods, such as sophisticated gradient descent versions, regularization techniques, or an optimal learning rate strategy.

\noindent
\textbf{Implications for Real-world Applications.} For practical applications, it is essential to consider both accuracy and processing speed. Although Defense-GAN offers reasonable accuracy, its extended processing time may limit its effectiveness for real-time use. In contrast, our approach achieves equilibrium, making it ideal for critical tasks of time-sensitiveness such as medical imaging and autonomous vehicles.

\noindent
\textbf{Direct Tensor Data Processing.} Within our investigation, we observed that saving and then reloading perturbed images notably improved the accuracy of image classifiers. However, these modifications could not be detected by the human eye. Several factors might account for this, such as the inherent volatility of adversarial attacks, potential loss of image data despite using lossless PNG formats during saving, and possible information loss due to normalization when loading images, among other unknown reasons. To ensure that our experimental results were not influenced by this effect, we opted to use the tensor data directly in the testing phase, avoiding the save and load steps. Our aim was to evaluate all defense methods, including ours, under the same conditions to maintain comparability of the results. Nonetheless, employing tensor data during testing could be a factor contributing to the poorer performance of other defense methods compared to their original reported outcomes.

\section{Conclusion}
This study reveals that certain defense mechanisms exhibit generalizability against adversarial attacks, notably the image reconstruction model grounded in the image-to-image translation approach. Additionally, we demonstrate that incorporating adversarial examples from various attack types during training outperforms using a single attack type. Our findings also indicate that a robust defense trained on six distinct adversarial attacks achieves an average restored classification accuracy comparable to an average of six individual attack-specific defenses. These results highlight the promise of defenses based on image-to-image translation for developing a comprehensive model that maintains stability when faced with unfamiliar attacks. This defense approach offers reduced training costs and the potential for broad application in practical scenarios. The next phase of this research involves developing an application that utilizes this adaptable model and evaluates its efficiency with printed adversarial example images.

\bibliographystyle{IEEEtran}
\bibliography{your_bib_file}

\newpage

\end{document}